\pdfoutput=1

\documentclass[11pt]{article}
\usepackage[dvipsnames]{xcolor}
\usepackage{amsmath} 
\usepackage{bm}
\usepackage{microtype}
\usepackage{hyperref}
\usepackage{url}
\usepackage{multirow}

\usepackage{tikz,tkz-tab}
\usepackage{colortbl}
\usepackage{booktabs}
\usepackage{graphicx}

\usepackage[final]{acl}

\usepackage{times}
\usepackage{latexsym}

\usepackage[T1]{fontenc}

\usepackage[utf8]{inputenc}

\usepackage{microtype}

\usepackage{inconsolata}

\usepackage{graphicx}
\graphicspath{ }
\title{Counterfactuals As a Means for Evaluating Faithfulness of Attribution Methods in Autoregressive Language Models}

\author{Sepehr Kamahi$^{1}$, Yadollah Yaghoobzadeh$^{1,2}$\\
        $^1$School of Electrical and Computer Engineering \\
  College of Engineering, University of Tehran, Tehran, Iran \\
     $^2$Tehran Institute for Advanced Studies, Khatam University, Tehran, Iran \\
   \texttt{sepehr.kamahi@ut.ac.ir},  \texttt{y.yaghoobzadeh@ut.ac.ir}}

\begin{document}
\maketitle
\begin{abstract}

Despite the widespread adoption of autoregressive language models, explainability evaluation research has predominantly focused on span infilling and masked language models. Evaluating the faithfulness of an explanation method—how accurately it explains the inner workings and decision-making of the model—is challenging because it is difficult to separate the model from its explanation. Most faithfulness evaluation techniques corrupt or remove input tokens deemed important by a particular attribution (feature importance) method and observe the resulting change in the model's output. However, for autoregressive language models, this approach creates out-of-distribution inputs due to their next-token prediction training objective. In this study, we propose a technique that leverages counterfactual generation to evaluate the faithfulness of attribution methods for autoregressive language models. Our technique generates fluent, in-distribution counterfactuals, making the evaluation protocol more reliable.
\end{abstract}

\section{Introduction}

Most modern NLP systems rely on autoregressive, transformer-based language models \citep{NEURIPS2020_1457c0d6,Touvron2023Llama2O,groeneveld2024olmo}. These models are inherently opaque, creating a strong need to understand their decision-making processes. As a result, explanation methods have become increasingly important in the field.

A widely-used approach for model explainability is attribution, also known as feature importance (FI) \citep{zhao2023explainabilitylargelanguagemodels}. Attribution methods aim to identify which input features contribute most to a model’s predictions, assigning a scalar value to each feature that reflects its relevance in the decision-making process. In typical NLP tasks, input features are often subwords or their combinations.

A key challenge in evaluating the faithfulness of attribution methods is that many existing techniques are designed for denoising or masked language models (MLMs) \citep{kobayashi-etal-2020-attention, kobayashi-etal-2021-incorporating, ferrando-etal-2022-measuring,modarressi-etal-2022-globenc,modarressi-etal-2023-decompx, mohebbi-etal-2023-quantifying}. Recent work on autoregressive models has primarily focused on the plausibility of attributions \citep{yin-neubig-2022-interpreting,ferrando-etal-2023-explaining}. While plausible (or persuasive) explanations might be the objective of the explainer, the core objective for the user is to truly understand the model’s decision-making process, rather than simply being convinced that the model's decisions are correct \citep{jacovi-goldberg-2021-aligning}.

Nearly all previous methods for faithfulness evaluation modify the input in some way, such as masking or removing important tokens based on the attribution results, and then measuring the impact on the model’s predictions. These methods tend to work well for MLMs, which are specifically trained for tasks like span or mask infilling. However, in the case of autoregressive models like GPT-2, which predict the next token, such modifications produce out-of-distribution (OOD) inputs. This raises a crucial question: are these evaluation methods truly assessing the informativeness of the selected tokens, or merely testing the model’s robustness to unnatural text and the artifacts introduced by testing modifications \citep{NIPS2019_9167}?
Moreover, the OOD nature of these inputs results in explanations that become socially misaligned \citep{NEURIPS2021_1def1713}. In other words, the expectations of users—who seek to understand which features are most relevant to the model’s decision—no longer align with the actual output of the attribution method. Instead, feature importance becomes influenced by the model’s priors rather than the learned features that truly drive predictions.

In this work, drawing inspiration from counterfactual generation—where the input is altered to flip the model’s output—we propose a new technique to evaluate the faithfulness of attribution methods in autoregressive language models. Specifically, we use counterfactual generators to modify the input by focusing on tokens highlighted by attribution methods, while ensuring that the altered input remains natural, fluent, and within the model’s original distribution. This ensures that any observed change in the model’s predictions is due to the modification of the important tokens, rather than an effect of OOD inputs.

We argue that if an attribution method enables a counterfactual generator to modify fewer tokens to change the model’s prediction, then it demonstrates a stronger understanding of the model’s inner workings, indicating higher faithfulness.
To validate our approach, we apply this faithfulness evaluation technique to several attribution methods—including gradient norm, gradient × input, erasure, KernelSHAP, and integrated gradients—within the context of next-word prediction for two language models: the fine-tuned Gemma-2b and the off-the-shelf Gemma-2b-instruct \citep{gemmateam2024gemma}.

Our contributions are as follows: (i) We introduce a novel faithfulness evaluation protocol that preserves the model’s input distribution, designed for attribution methods in autoregressive language models. (ii) We apply this protocol to evaluate and rank widely-used attribution methods, showcasing differences in sensitivity between fine-tuned and off-the-shelf models when handling OOD data and proposing a solution.\footnote{The code is available at \url{ https://github.com/Sepehr-Kamahi/faith}}

\section{Related work}

\textbf{Evaluating Explanations.} One line of work uses another method as a proxy for evaluating an explanation. For instance, \citet{abnar-zuidema-2020-quantifying} assess explanations by comparing them with gradient. Although \citet{wiegreffe-pinter-2019-attention} caution that gradients should not be considered ideal or as the ``ground truth,'' they nonetheless utilize gradients as a proxy for the model's intrinsic semantics.

Most current metrics for evaluating faithfulness involve either removing important tokens or retraining the model using only those identified as important by attribution methods \citep{chan-etal-2022-comparative}.  Importantly, the trustworthiness of explanations is both task- and model-dependent \citep{bastings-etal-2022-will}, and different attribution methods frequently produce inconsistent results \citep{neely2022song}. As a result, it is not justifiable to treat any single explanation method as a universal standard across all contexts.

In their work, \citet{deyoung-etal-2020-eraser} introduce two key concepts: comprehensiveness (whether the important tokens identified are the only ones necessary for making a prediction) and sufficiency (whether these important tokens alone are enough to make the prediction). \citet{carton-etal-2020-evaluating} build on this by proposing normalized versions of these concepts, comparing comprehensiveness and sufficiency to the null difference—the performance of an empty input (for sufficiency) or a full input (for comprehensiveness). However, it remains unclear whether these corruption techniques evaluate the informativeness of the corrupted tokens or merely the robustness of the model to unnatural inputs and artifacts introduced during evaluation.

Further, \citet{han-etal-2020-explaining} and \citet{jain-etal-2020-learning} frame attribution methods as either faithful or unfaithful, with no consideration for degrees of faithfulness. They describe attribution methods that are ``faithful by construction.'' In contrast, other researchers propose that faithfulness exists on a spectrum and suggest evaluating the ``degree of faithfulness'' of explanation methods \citep{jacovi-goldberg-2020-towards}. Our approach aligns with this view, as we aim to find explanation methods that are sufficiently faithful for autoregressive models.

\citet{atanasova-etal-2023-faithfulness} evaluate the faithfulness of natural language explanations using counterfactuals, applying techniques from \citet{ross-etal-2021-explaining} to assess how well explanations align with the model’s decision-making. This line of work offers valuable insights into the use of counterfactuals, which we build upon for evaluating attribution methods in language models.

\textbf{The OOD Problem in Explainability.}  
The issue of OOD inputs in explainability has been raised by several works. \citet{NIPS2019_9167} and \citet{vafa-etal-2021-rationales} suggest retraining or fine-tuning the model using partially erased inputs to align training and evaluation distributions. However, this process can be computationally expensive and is not always practical. An alternative approach by \citet{kim-etal-2020-interpretation} aims to ensure that the explanation remains in-distribution to mitigate OOD problems. Our work addresses this concern by preserving the input distribution during faithfulness evaluation, particularly for autoregressive models.

\textbf{Feature Importance (Attribution).}
Attributions, or feature importance scores, are local explanations that assign a score to each input feature—typically token embeddings in NLP tasks—indicating how crucial that feature is to the model’s prediction. Attribution methods can be categorized into four types:
i) Perturbation-based methods, which alter or mask input features to assess their importance by observing changes in the model’s output \citep{li-etal-2016-visualizing,li2017understanding,feng-etal-2018-pathologies,wu-etal-2020-perturbed}.
ii) Gradient-based methods, which calculate the derivative of the model’s output with respect to each input to measure the influence of each feature \citep{mohebbi-etal-2021-exploring,Kindermans2019,pmlr-v70-sundararajan17a,pmlr-v162-lundstrom22a,enguehard-2023-sequential,sanyal-ren-2021-discretized,sikdar-etal-2021-integrated}.
iii) Surrogate-based methods, which explain a complex black-box model using a simpler, interpretable model \citep{ribeiro-etal-2016-trust,NIPS2017_8a20a862,kokalj-etal-2021-bert}.
iv) Decomposition-based methods, which break down the overall importance score into linear contributions from the input features \citep{Montavon2019,voita-etal-2021-analyzing,9577970,modarressi-etal-2022-globenc,ferrando-etal-2022-towards}.


\section{Our method } \label{methods}

Our faithfulness evaluation protocol consists of two models: a counterfactual generator and a predictor model. Our goal is to assess the faithfulness of attribution methods used for the predictor model. Given the large output space of autoregressive language models (LMs)—often encompassing thousands of vocabulary items—analyzing the entire output space provides limited insight. To address this, we adopt contrastive explanations from \citet{yin-neubig-2022-interpreting}, which evaluate the attribution of input tokens for a contrastive decision. Specifically, contrastive attributions identify the most influential tokens that led the model to predict the target label $y_t$ over an alternative foil $y_f$. These important tokens are then modified by a separate editor model to generate counterfactuals, i.e., examples that shift the prediction of the original model towards the foil.

\begin{figure*}[t]
\begin{center}

\includegraphics[width=0.98\linewidth]{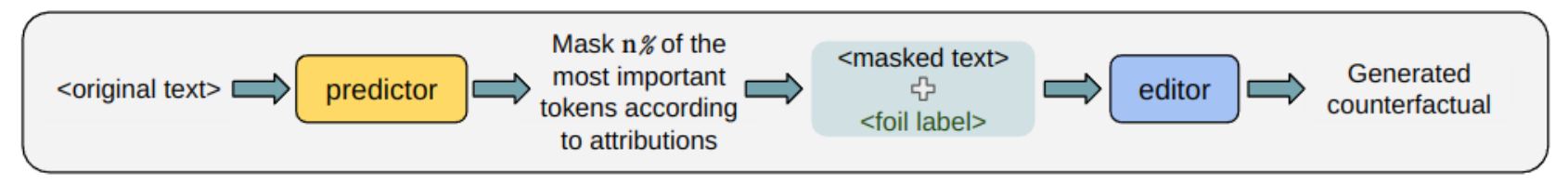}
\end{center}
\caption{Our process of generating counterfactuals for evaluating attribution methods. The predictor (an LM) generates a label for the given text, and an attribution method specifies the most important tokens. We mask the top n\% of them and ask an editor (another LM) to change the label of the input text by filling in the masked tokens. If the attribution method is more faithful, then the required n\% should be lower.}
\label{protocol_simple}
\end{figure*}

Our evaluation protocol for attributions involves two key phases. In the first phase, we create an editor capable of generating counterfactuals. The second phase involves using both the editor and predictor to determine the minimum percentage of tokens the editor must change to flip the predictor's prediction. Figure \ref{protocol_simple} illustrates the second phase of this process.

\begin{figure*}[t]
\begin{center}

\includegraphics[width=0.98\linewidth]{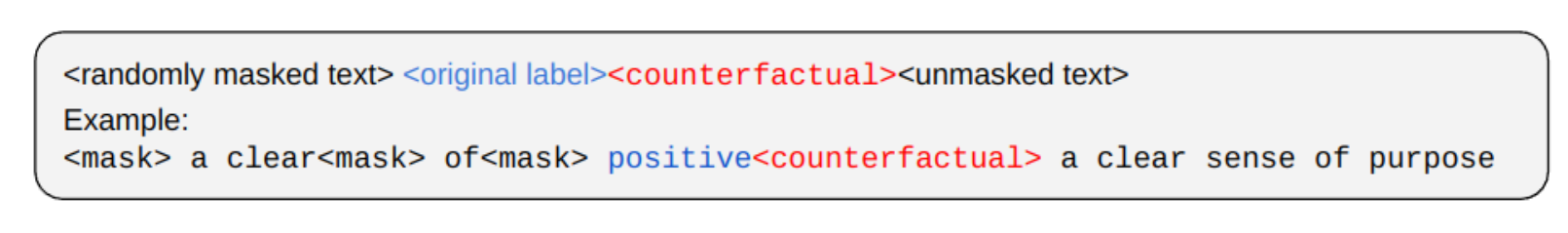}
\end{center}
\caption{Creation of training examples for fine-tuning the counterfactual generator, and one given sample.}
\label{cfg_ft}
\end{figure*}

To construct the editor, we fine-tune an autoregressive language model for the task of counterfactual generation. During fine-tuning, we expand the model's embedding space and tokenizer by adding two special tokens: \textless{}mask\textgreater' and \textless{}counterfactual\textgreater'. Inspired by \citet{wu-etal-2021-polyjuice} and \citet{donahue-etal-2020-enabling}, we generate training examples by randomly masking between 5\% and 50\% of the tokens in each example. To each example, we append its label (e.g., positive or negative for the SST-2 dataset), the `\textless{}counterfactual\textgreater' token, and the original unmasked example. Figure \ref{cfg_ft} demonstrates this process of constructing the training examples.

\begin{figure}[t]
\begin{center}

\includegraphics[width=1.0\linewidth]{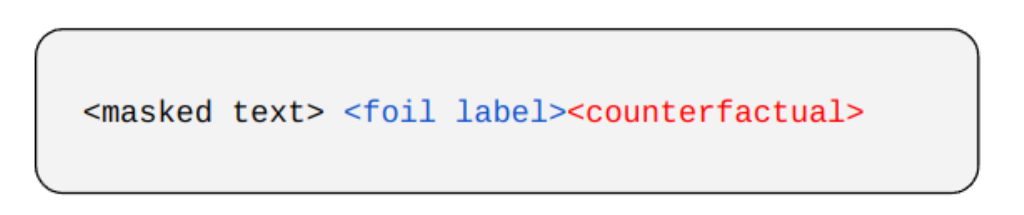}
\end{center}
\caption{Prompting techniques used for counterfactual generation in the second phase.}
\label{prompt_it}
\end{figure}

In the second phase of attribution evaluation, we input a sentence into the predictor model and apply an attribution method to identify the most important tokens influencing its prediction. Starting with the top 10\% most important tokens, we replace these tokens with `\textless{}mask\textgreater'. The masked sentence, along with the foil label (the label associated with the second-highest logit), is then passed to the \textit{editor}, which generates a counterfactual sentence intended to flip the predictor model’s output. If the prediction does not flip, we incrementally increase the masking by 10\% until either the prediction is flipped or we reach a masking threshold of 50\%. This process is outlined in Figure \ref{protocol_simple}, and the prompting technique for counterfactual generation is detailed in Figure \ref{prompt_it}. The attribution method that enables the editor to flip the prediction with the fewest changes is deemed to provide the most faithful representation of the predictor model's decision-making process.

\section{Experimental Setup}

\subsection{Datasets}

We use three datasets for evaluating faithfulness: SST-2 \citep{socher-etal-2013-recursive} and IMDB \citep{maas-etal-2011-learning}, which are both binary classification datasets, and AG-News \citep{NIPS2015_250cf8b5}, a four-class classification dataset.

Faithfulness evaluation datasets should not have gold attribution labels because we do not want human intuition to influence the evaluation. Instead, we aim to understand how the model makes predictions \citep{jacovi-goldberg-2020-towards}.

\subsection{Models}
\subsubsection{Editor Models}
For the editor model, our method is similar to \citet{wu-etal-2021-polyjuice}, which uses GPT-2, a decoder-only causal model, for generating counterfactuals. We extend this by using three more modern decoder-only models: GPT-J-6B \citep{gpt-j}, which we refer to as "gptj," and two sizes of Pythia: Pythia-1.4B (pythia1) and Pythia-2.8B (pythia2) \cite{biderman2023pythia}. We fine-tune these models following the process described in Section \ref{methods}. The pythia1 model is fully fine-tuned, while the other two (gptj and pythia2) are fine-tuned using Low-Rank Adaptation (LoRA) \citep{hu2022lora}. All models are trained for 8 epochs using dynamic masking \citep{liu2019roberta}, meaning each example is masked differently in each epoch.
\subsubsection{Predictor Models}
We use Gemma-2b \citep{gemmateam2024gemma} as the predictor model. We fine-tune the raw language model for the three datasets (referred to as gemma-ft) using Low-Rank Adaptation (LoRA). Additionally, we employ an off-the-shelf instruct-tuned version (gemma-it) for zero-shot evaluation. We then conduct a detailed comparison between these two versions—fine-tuned (gemma-ft) and non-fine-tuned (gemma-it)—to assess their differences in attribution evaluation.
\subsection{Attribution Methods}
Here we detail the six widely used attribution methods employed in our study. We use all attribution methods in a contrastive way \citep{yin-neubig-2022-interpreting}. Contrastive attributions measure which features from the input make the foil token $y_f$ more likely and the target token $y_t$ less likely. We denote contrastive, target, and foil attributions by $S^C$, $S^t$, and $S^f$ respectively:
\vspace{-1.0ex}
\begin{equation}
    S^C  = S^t - S^f
    \label{eq:contrastive_definition}
\vspace{-1.0ex}
\end{equation}
We use the implementation of these attribution methods provided by \citet{yin-neubig-2022-interpreting} (for Gradient $\times$ input, gradient norm and erasure) and by Captum \citep{miglani-etal-2023-using} (for KernelSHAP and Integrated Gradient).

\subsubsection{Gradient Norm}
We can calculate attributions based on the norm of the gradient of the model's prediction with respect to the input $x$ \citep{simonyan2013deep, li-etal-2016-visualizing}.The gradient with respect to feature $x_i$ is given by:
$$
g(x_i) = \nabla_{x_i} q(y_t | x)
$$
Where $q(y_t | x)$ is the model output for token $y_t$ given the input $x$.
The contrastive gradient:
$$
g^C(x_i) = \nabla_{x_i} \left( q(y_t | \bm{x}) - q(y_f | \bm{x}) \right)
$$ 
We will use both norm one (gradnorm1) and norm two (gradnorm2):
$$
S_{GN1}^C(x_i) = ||g^C(x_i)||_{L1}
$$
$$
S_{GN2}^C(x_i) = ||g^C(x_i)||_{L2}
$$

\subsubsection{Gradient $\times$ Input}

In gradient $\times$ input (gradinp) method \citep{shrikumar2016not, denil2014extraction}, we compute the dot product of the gradient and the input token embedding $x_i$:
$$
S_{GI}(x_i) = g(x_i) \cdot x_i
$$
By multiplying the gradient by the input embedding, we also account for how much each token is expressed in the attribution score. The Contrastive Gradient $\times$ Input is:
$$
S^C_{GI}(x_i) = g^C(x_i) \cdot x_i
$$

\subsubsection{Erasure}
Erasure-based methods measure the importance of each token by erasing it and observing the effect on the model output \citep{li2017understanding}. This is achieved by taking the difference between the model output with the full input $x$ and the model output with the input where token $x_i$ is zeroed out, denoted as $x_{\neg i}$:

$$S_E^t(x_i) = q(y_t|x) - q(y_t|x_{\neg i})$$

\noindent For the contrastive case, $S^C_E(x_i)$ becomes:
$$
(q(y_t|x) - q(y_t|x_{\neg i})) - (q(y_f|x) - q(y_f|x_{\neg i}))
$$
\subsubsection{KernelSHAP}
KernelSHAP \citep{NIPS2017_8a20a862} explains the prediction of a classifier $q$ by learning a linear model $\phi$ locally around each prediction. The objective function of KernelSHAP constructs an explanation that approximates the behavior of $q$ accurately in the neighborhood of $x$. More important features have higher weights in this linear model $\phi$. Let $Z$ be a set of $N$ randomly sampled perturbations around $x$:

\vspace{-1.0ex}
\begin{equation}
    S_{\phi}^t  = \underset{\phi}{\arg\min}\sum_{z \in Z}[q(y_t|z)-\phi^Tz]^2\pi_x(z)
    \label{eq:kernelshap}
\vspace{-1.0ex}
\end{equation}

\noindent KernelSHAP uses a kernel $\pi_x$ that satisfies certain principles when input features are considered agents of a cooperative game in game theory. We use equation \ref{eq:kernelshap} in a contrastive way. First we normalize $S_{\phi}^t$ and $S_{\phi}^f$ by dividing by their $L2$ norm and then subtracting:

\vspace{-1.0ex}
\begin{equation}
    S_{\phi}^C = \frac{S_{\phi}^t}{||S_{\phi}^t||} - \frac{S_{\phi}^f}{||S_{\phi}^f||}
    \label{eq:kernelshap_contrastive}
\vspace{-1.0ex}
\end{equation}

\subsubsection{Integrated Gradients}
Integrated Gradients (\textsc{IG}) \citep{pmlr-v70-sundararajan17a} is a gradient-based method which addresses the problem of saturation: gradients may get close to zero for a well-fitted function. IG requires a baseline $\mathbf{b}$ as a way of contrasting the given input with the absence of information.  
For input $i$, we compute:
\begin{equation}
    S^t_{IG}(x_i) = \frac{1}{m}  \sum_{k=1}^m \nabla_{x_i} q\Big(y_t\Big|b+ \frac{k}{m}  (x-b )\Big) \!\cdot\!  (x_i \!-\! b_i) \!
\end{equation}
\noindent That is, we average over $m$ gradients, with the inputs to $q$ being linearly interpolated between the baseline $b$ and the original input $x$ in $m$ steps. We then take the dot product of that averaged gradient with the input embedding $\mathbf{x}_i$ minus the baseline.

We use a zero vector baseline \citep{mudrakarta-etal-2018-model} and five steps. The contrastive case becomes:

\vspace{-1.0ex}
\begin{equation}
    S_{IG}^C = \frac{S_{IG}^t}{||S_{IG}^t||} - \frac{S_{IG}^f}{||S_{IG}^f||}
    \label{eq:IG_contrastive}
\vspace{-1.0ex}
\end{equation}

\label{gen_inst}

\begin{table*}[ht]

\begin{center}
\begin{tabular}{m{3cm}||cc|cc|cc}
\toprule
\multirow{2}{0pt}{\bf Editor} &\multicolumn{2}{c}{\bf gradnorm1} &\multicolumn{2}{c}{\bf Erasure} &\multicolumn{2}{c}{\bf KernelSHAP}\\

  &{\bf gemma-ft} & {\bf gemma-it } &{\bf gemma-ft} & {\bf gemma-it } &{\bf gemma-ft} & {\bf gemma-it }\\
\midrule
pythia1 (ours) & 1.1  & 1.4    & 1.3   & 1.6 & 0.7   & 1.7\\
pythia2 (ours) & 0.4   & 2.6  & 0.9  & 1.3   & 0.8   & 2.3\\
gptj (ours)    & 0.7 & 8.3  & 2.0    & 10.9        & 0.9   & 6.4\\
erase          & 0.3     & \colorbox{Yellow}{19.9}      & 2.3         & \colorbox{Yellow}{32.8}        & 0.6   & \colorbox{Yellow}{81.4}\\
unk            & 0.6        & \colorbox{Yellow}{97.5}      & 1.8         & \colorbox{Yellow}{97.3}        & 1.3   &  \colorbox{Yellow}{99.8}\\   
mask          &   0.0     & \colorbox{Yellow}{94.8}      & 0.5         & \colorbox{Yellow}{93.3}        & 0.0   &  \colorbox{Yellow}{98.5}\\   
att-zero      & 0.1        & \colorbox{Yellow}{80.9}      & 0.1         & \colorbox{Yellow}{62.6}        & 0   &  \colorbox{Yellow}{74.1}\\    
\bottomrule
\end{tabular}
\end{center}
\caption{OOD percentage when our counterfactual editor models generate samples, compared to other replacement methods (erase, unk, mask, and att-zero methods). This represents the percentage of corrupted examples that fall outside the 99th percentile of the NLL of the original sentences in the SST-2 dataset (lower is better). Scenarios with very high OOD percentages are highlighted.}
\label{OODs}
\end{table*}

\begin{table*}[t]
\begin{center}
\begin{tabular}{m{2cm}||ccc|ccc|ccc}
\toprule
\multirow{2}{0pt}{\bf Attribution method} &\multicolumn{3}{c}{\bf SST-2} &\multicolumn{3}{c}{\bf IMDB} &\multicolumn{3}{c}{\bf AG-News}\\

  &{\bf pythia1} & {\bf pythia2 } &{\bf gptj} &{\bf pythia1} & {\bf pythia2 }&{\bf gptj}&{\bf pythia1} & {\bf pythia2 }&{\bf gptj}\\
\midrule
gradnorm1   &33.5  &{\bf 34.8} & {\bf32.2}& {\bf29.1} &{\bf 30.3} & {\bf32.4} &42.9 &45.1  & 44.0\\

gradnorm2 &{\bf33.4} &35.6    & 32.6 &31.0    &30.5   &  {\bf32.4}   &42.6  & 44.4 & 43.9\\

gradinp      &40.5   &41.8    &40.8 &36.1    &36.3     &  36.5 &43.1   &44.6  & {\bf42.2}\\

erasure      &35.5   &36.6     &33.4 &32.7    &32.7  & 34.4 &{\bf 42.0}   &{\bf 42.7} &43.0\\

IG           &45.7   &45.8     & 43.7&43.3    &44.3     &42.5 &43.8     &46.7 & 44.0\\

KernelSHAP   &44.1   &45.9     &44.9 &44.0     &43.3     & 44.2 &44.0     &46.5  & 44.3\\

Random       &44.6   &46.0     & 44.3 &43.8    &42.7      &43.2 &44.0     &46.0   & 44.0\\

\bottomrule
\end{tabular}
\end{center}
\caption{The mean percentage of tokens needed to be masked to achieve flipping Gemma-ft's label or reaching 50 percent masking in 200 examples from evaluation split of SST-2, IMDB, and AG-News datasets (lower is better). pythia1, pythia2, and gptj models are used to fill the masks and generate counterfactuals.}
\label{mask-percent-ft}
\end{table*}

\begin{table*}[ht]
\begin{center}
\begin{tabular}{m{2cm}||ccc|ccc|ccc}
\toprule
\multirow{2}{0pt}{\bf Attribution method} &\multicolumn{3}{c}{\bf SST-2} &\multicolumn{3}{c}{\bf IMDB} &\multicolumn{3}{c}{\bf AG-News}\\

  &{\bf pythia1} & {\bf pythia2 } &{\bf gptj} &{\bf pythia1} & {\bf pythia2 }&{\bf gptj}&{\bf pythia1} & {\bf pythia2 }&{\bf gptj}\\
\midrule
gradnorm1    &41.4     &42.0 & {\bf40.3}    &42.1      &42.3   & 44.0 &46.6     &46.8 & 40.0 \\
gradnorm2    &41.5     &42.2  & 40.6  &42.3      &42.6   & 43.6 &46.6     &46.9  & 39.9\\
gradinp      &42.9     &43.4   & 43.2 &{\bf41.2}      &{\bf41.4}    & {\bf42.2} &45.8    &45.3 & 39.2\\
erasure      &{\bf40.8}     &{\bf41.5} & 41.0 &43.1 &42.8  &43.8 &45.0     &45.5  & 39.5\\
IG          & 44.7     &44.0  &  45.4 &43.4     &43.2    & 44.0 &45.7      &45.3   & 38.6\\
KernelSHAP   &43.6     &43.5   & 44.0 &43.7   &  42.9   &43.9 &46.1   &45.3   &{\bf37.2} \\
Random       &44.8     &44.7  &  44.3 &45.4     &46.1    &  45.8 &{\bf44.9}     &{\bf45.2}   & 39.2\\
\bottomrule
\end{tabular}
\end{center}
\caption{The mean percentage of tokens needed to be masked to achieve flipping Gemma-it's label or reaching 50 percent masking in 200 examples from evaluation split of SST-2, IMDB, and AG-News datasets (lower is better). pythia1, pythia2, and gptj models are used to fill the masks and generate counterfactuals.}

\label{mask-percent-it}
\end{table*}

\section{Results and Discussion} \label{res}
This section begins by addressing the out-of-distribution (OOD) problem, emphasizing the importance of using counterfactual generators over methods like erasing or replacing important tokens with uninformative ones. First, we demonstrate that counterfactual generators produce in-distribution text for both fine-tuned and instruct-tuned models by employing OOD detection techniques focused on background shift, using negative log-likelihood (NLL) as a metric. We find that fine-tuned models are less sensitive to input corruptions, whereas instruct-tuned models classify altered inputs as OOD unless counterfactual generators are used. Second, we show that the faithfulness rankings of attribution methods remain consistent when using counterfactual generators, but differ significantly with other replacement methods, with Spearman's rank correlation used for evaluation. Lastly, in Subsection \ref{FI_sub}, we analyze feature importance methods by examining the average masking percentage required to flip model predictions. The results indicate that simpler attribution methods perform well for fine-tuned models across different datasets, but are generally less effective for instruct-tuned models, highlighting challenges in applying attribution methods to general-purpose models.
\subsection{The Out-of-Distribution Problem} \label{ood_sub}

\paragraph{Why should we use counterfactuals instead of erasing important tokens or replacing them with unimportant ones?}

First, we demonstrate that our counterfactual generators produce in-distribution text for the predictor. Second, we show that the faithfulness rankings of attribution methods are consistent when using a counterfactual generator for token replacement, but these rankings differ significantly when other replacement methods are applied.

To achieve our first goal—demonstrating that the generated counterfactuals remain in-distribution—we employ an OOD detection technique to measure the percentage of generated inputs classified as OOD. Prominent OOD detection methods use a threshold, where any input exceeding this threshold is considered OOD \citep{chen-etal-2023-fine}. For each dataset, we compute this threshold by measuring the negative log-likelihood (NLL) of 200 original examples using different predictors (fine-tuned and instruct-tuned) and set the 99th percentile of these NLLs as the OOD threshold. We use NLL for detecting OOD instances because the type of shift we focus on is background shift. OOD data can be categorized as either semantic or background shift \citep{arora-etal-2021-types}. 
Semantic features have a strong correlation with the label, and a semantic shift occurs when we encounter unseen classes at test time. In contrast, background shift involves population-level statistics that do not depend on the label and pertain more to the style of the text.

In our evaluation of faithfulness through input corruption, we do not introduce new labels or classes but instead alter the style of the text. Therefore, we focus on detecting background shift. There are two main approaches to OOD detection: calibration-based methods and density estimation methods. Density estimation, such as perplexity (PPL), typically outperforms calibration-based methods when dealing with background shifts, whereas the opposite is true for semantic shifts. Since NLL is closely related to PPL, it is used in our experiments.

An attribution method highlights important tokens, and we replace those tokens using four different approaches: (i) our method of using an editor for token replacement, (ii) replacing tokens with semantically unimportant ones (such as the \textless{}unk\textgreater{} or \textless{}mask\textgreater{} tokens), (iii) erasing the tokens, and (iv) zeroing out the attention mask for important tokens without altering the text itself (att-zero).

The baseline methods (ii) through (iv) are similar to those used in prior work \citep{NEURIPS2021_1def1713}. Table~\ref{OODs} shows that, for both fine-tuned and instruct-tuned predictors, the counterfactual generator consistently produces in-distribution text. We present results for the SST-2 dataset using three different attribution methods, while results for other methods and datasets are included in Appendix~\ref{A}. Each value in Table~\ref{OODs} represents the average over five levels of replacement (ranging from 10\% to 50\%) and 200 examples from evaluation sets.

\citet{chen-etal-2023-fine} show that fine-tuning makes models less sensitive to non-semantic shifts. Fine-tuning reduces task-agnostic, pre-trained knowledge of general linguistic properties that are crucial for detecting such shifts. Our findings align with this observation. When a model is fine-tuned for a specific classification task—such as sentiment analysis on SST-2—it becomes optimized to predict labels with high confidence based on training data. As a result, fine-tuned models are more resilient to input corruptions, regardless of the replacement method. As Table~\ref{OODs} illustrates, under the Gemma-ft columns, the percentage of OOD examples is close to zero for the fine-tuned predictor, regardless of the replacement strategy used.

In contrast, Gemma-it, an off-the-shelf model not fine-tuned for a specific dataset, behaves differently. When important tokens are replaced with semantically neutral tokens (e.g., \textless{}unk\textgreater{} or \textless{}mask\textgreater{}), erased, or subjected to attention masking without textual alteration, the Gemma-it predictor frequently categorizes these modified inputs as OOD. This difference in behavior between fine-tuned and off-the-shelf models highlights the impact of task-specific optimization on a model's sensitivity to input perturbations. However, when using the counterfactual generator to modify inputs, the modified examples remain in-distribution even for the instruct-tuned predictor. This demonstrates that counterfactual generators are useful for evaluating the faithfulness of attribution methods, especially when using off-the-shelf models where modification without creating OOD inputs is essential.

To achieve our second goal—demonstrating that faithfulness rankings of attribution methods remain consistent when using a counterfactual generator, and inconsistent with other replacement methods—we use Spearman’s rank correlation, as in prior studies \citep{pmlr-v162-rong22a}. For each example, we rank attribution methods based on the percentage of the mask required to flip the label. We then compute the correlation between these rankings across seven replacement methods (three editors, Erase, \textless{}unk\textgreater{}, \textless{}mask\textgreater{}, and att-zero), averaging the results over 200 examples.

Figure~\ref{corr_sst} presents this analysis for the SST-2 dataset. Similar results for other datasets are shown in Appendix~\ref{B}. The top matrix in Figure~\ref{corr_sst} shows average correlation results for the fine-tuned predictor, where all replacement methods exhibit high average correlations with one another. The middle matrix depicts the correlations for the off-the-shelf instruct-tuned model. In this case, only when using a counterfactual generator do the rankings show high correlation, while other replacement methods produce low correlations. This is likely due to OOD inputs being created by non-counterfactual replacement methods.

The bottom matrix in Figure~\ref{corr_sst} illustrates the difference between the first and second matrices. It shows that the correlation difference between fine-tuned and instruct-tuned predictors is negligible when editors are used for token replacement. However, significant differences arise when other methods (i.e., \textless{}unk\textgreater{}/Erase/\textless{}mask\textgreater{}/att-zero) are employed, underscoring the importance of avoiding OOD inputs when evaluating explanations for off-the-shelf models.

\begin{figure}[h!]

\begin{center}

\includegraphics[width=1.02\linewidth]{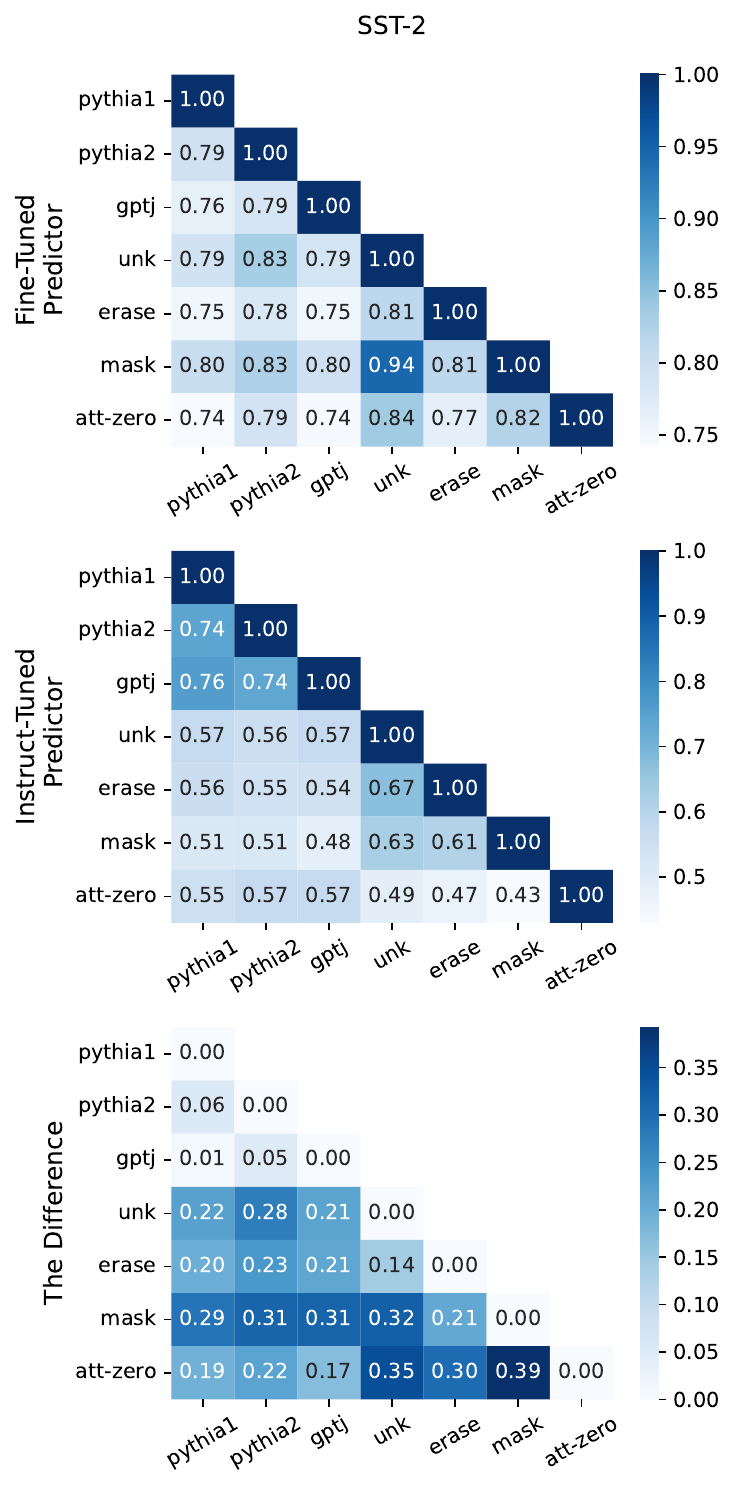}
\end{center}
\caption{The top matrix presents the average correlation of attribution ranks for the fine-tuned predictor. The middle matrix shows the average correlation of attribution ranks when using an off-the-shelf instruct-tuned predictor. The bottom matrix illustrates the difference between the fine-tuned and instruct-tuned models, indicating that when editors are used as the replacement method, the difference in correlation is near zero. In contrast, using other replacement methods (i.e., \textless{}unk\textgreater{}, erase, \textless{}mask\textgreater{}, att-zero) results in significant inconsistencies between the two predictor types, likely due to the creation of out-of-distribution (OOD) text for the instruct-tuned model.}
\label{corr_sst}
\end{figure}

\subsection{Analysis of Feature Importance Methods} \label{FI_sub}

Tables~\ref{mask-percent-ft} and \ref{mask-percent-it} present the average masking percentage required (i.e., the average percentage of tokens that the counterfactual generator needs to modify) to flip the prediction for fine-tuned and instruct-tuned predictor models, respectively. This masking percentage is strongly correlated with the flip rate—the proportion of labels the counterfactual generator successfully flips by altering the identified important tokens. In Appendix~\ref{C}, we provide a detailed analysis of the flip rate for both fine-tuned and instruct-tuned models. Attribution methods that require less masking to flip a label tend to also have a higher flip rate, indicating greater faithfulness.

For the fine-tuned predictor (Table~\ref{mask-percent-ft}), gradient norm-based methods consistently outperform others on the SST-2 and IMDB datasets. However, for AG-News, the Erasure method consistently performs the best or near the best. These results suggest that simpler attribution methods, such as gradnorm1, gradnorm2, and Erasure, provide superior performance across different editors and datasets.

For the instruct-tuned predictor (Table~\ref{mask-percent-it}), the Erasure method achieves the best results for the SST-2 dataset, while the gradinp method performs best on the IMDB dataset. Interestingly, for the AG-News dataset, no attribution method consistently outperforms random selection, highlighting the challenges of applying attribution methods to general-purpose models. These findings emphasize that attribution methods are generally less effective when the model is not fine-tuned for a specific task, suggesting the need for careful consideration when applying these methods to pretrained or instruct-tuned models.

\section{Conclusion}
In this work, we introduced a faithfulness evaluation protocol based on counterfactual generation to assess the performance of attribution methods. We demonstrated that the efficacy of these methods differs significantly between models that are fine-tuned on a specific dataset and those that are instruct-tuned and used off-the-shelf. Our results showed that counterfactual generators are highly effective for evaluating feature attribution, as they consistently produce in-distribution text for the predictor model. This ensures a clearer separation between the evaluation of the model itself and the attribution method, with minimal influence from out-of-distribution examples.

Moreover, our findings revealed strong consistency across different counterfactual generators, while other token replacement methods lacked this consistency. This underscores the importance of maintaining in-distribution inputs, especially when evaluating attribution methods for off-the-shelf models. Lastly, we applied our protocol to compare various attribution methods, providing insights into their relative effectiveness across different model configurations.

\section{Limitations}
Our work has several limitations. First, it relies on the generation of counterfactuals, which necessitates the use of a robust generative model. This process, particularly for long sequences, can be computationally expensive and may limit the scalability of our approach. Second, there is a risk that the counterfactual generator may unintentionally exploit artifacts or shortcuts used by the predictor model to flip the label, which could undermine the broader applicability of our method. Finally, our protocol has been tested exclusively on classification tasks; extending it to other tasks, such as machine translation or question answering, remains an area for future exploration.

\label{others}

\bibliography{custom}

\appendix
\section{} \label{A}

Table \ref{OODs_2} is the OOD percentages for other attribution methods in SST-2 dataset that were not in Table \ref{OODs}. Tables \ref{OODs_news1} and \ref{OODs_nnews2} show OOD percentages in AG-News dataset.

\begin{table*}[h]

\begin{center}
\begin{tabular}{m{3cm}||cc|cc|cc}
\toprule
\multirow{2}{0pt}{\bf Editor} &\multicolumn{2}{c}{\bf gradnorm2} &\multicolumn{2}{c}{\bf gradinp} &\multicolumn{2}{c}{\bf integrated gradient}\\

  &{\bf gemma-ft} & {\bf gemma-it } &{\bf gemma-ft} & {\bf gemma-it } &{\bf gemma-ft} & {\bf gemma-it } \\
\midrule
pythia1 (ours) & 1.4  & 1.2    & 1.3   & 1.5 & 1.1   & 0.7\\
pythia2 (ours) & 0.4   & 2.6  & 0.8  & 1.6   & 1.0   & 0.8\\
gptj (ours)    & 0.8 & 8.2  & 2.4    & 5.2        & 0.4   & 8.7\\
erase          & 0.5     & \colorbox{Yellow}{21.9}      & 1.1         & \colorbox{Yellow}{83.8}        & 1.4   & \colorbox{Yellow}{77.4}\\
unk            & 0.6        & \colorbox{Yellow}{98.4}      & 1.6         & \colorbox{Yellow}{100.0}        & 3.4   &  \colorbox{Yellow}{99.3}\\   
mask          &   0.0     & \colorbox{Yellow}{95.8}      & 0.2         & \colorbox{Yellow}{99.5}        & 0.4   &  \colorbox{Yellow}{97.8}\\   
att-zero      & 0.1        & \colorbox{Yellow}{80.8}      & 0.0         & \colorbox{Yellow}{70.6}        & 0   &  \colorbox{Yellow}{65.8}\\    
\bottomrule
\end{tabular}
\end{center}
\caption{OOD percentage when our counterfactual editor models generate samples, compared to other replacement methods (erase, unk, mask, and att-zero methods). This is the percentage of corrupted examples that are out of the 99th percentile of the NLL of the original sentences in SST-2 dataset (lower is better). The scenarios with very high numbers of OODs are highlighted.}
\label{OODs_2}
\end{table*}

\begin{table*}[h]

\begin{center}
\begin{tabular}{m{3cm}||cc|cc|cc}
\toprule
\multirow{2}{0pt}{\bf Editor} &\multicolumn{2}{c}{\bf gradnorm1} &\multicolumn{2}{c}{\bf Erasure} &\multicolumn{2}{c}{\bf KernelSHAP}\\

  &{\bf gemma-ft} & {\bf gemma-it } &{\bf gemma-ft} & {\bf gemma-it } &{\bf gemma-ft} & {\bf gemma-it }\\
\midrule
pythia1 (ours) & 2.4  & 5.0    & 4.1   & 4.9 & 1.6   & 4.5 \\
pythia2 (ours) & 2.4   & 5.2  & 3.1  & 4.9   & 2.0   & 6.9\\
gptj (ours)    & 1.0 & 5.0  & 1.5    & 4.9        & 0.9   & 7.8\\
erase          & 3.5     & \colorbox{Yellow}{11.5}      & 8.8         & \colorbox{Yellow}{46.3}        & 2.3   & \colorbox{Yellow}{74.9}\\
unk            & 1.0        & \colorbox{Yellow}{98.2}      & 3.8         & \colorbox{Yellow}{99.7}        & 1.3   &  \colorbox{Yellow}{99.9}\\   
mask          &   0.7     & \colorbox{Yellow}{85.7}      & 5.2         & \colorbox{Yellow}{96.7}        & 0.5   &  \colorbox{Yellow}{98.2}\\   
att-zero      & 4.9        & \colorbox{Yellow}{79.1}      & 3.2         & \colorbox{Yellow}{57.4}        & 0.8   &  \colorbox{Yellow}{62.6}\\    
\bottomrule
\end{tabular}
\end{center}
\caption{OOD percentage when our counterfactual editor models generate samples, compared to other replacement methods (erase, unk, mask, and att-zero methods). This is the percentage of corrupted examples that are out of the 99th percentile of the NLL of the original sentences in AG-News dataset (lower is better). The scenarios with very high numbers of OODs are highlighted.}
\label{OODs_news1}
\end{table*}

\begin{table*}[h]

\begin{center}
\begin{tabular}{m{3cm}||cc|cc|cc}
\toprule
\multirow{2}{0pt}{\bf Editor} &\multicolumn{2}{c}{\bf gradnorm2} &\multicolumn{2}{c}{\bf gradinp} &\multicolumn{2}{c}{\bf integrated gradient}\\

  &{\bf gemma-ft} & {\bf gemma-it } &{\bf gemma-ft} & {\bf gemma-it } &{\bf gemma-ft} & {\bf gemma-it } \\
\midrule
pythia1 (ours) & 2.2  & 5.0    & 1.4   & 4.5 & 1.6   & 4.5\\
pythia2 (ours) &  2.4  & 5.2  & 2.3  & 5.5   & 1.3   & 5.9 \\
gptj (ours)    & 1.2 & 5.0  & 2.0    & 5.1        & 1.2   & 6.4\\
erase        & 3.3     & \colorbox{Yellow}{11.6}      & 2.6         & \colorbox{Yellow}{60.3}        & 1.9   & \colorbox{Yellow}{55.5}\\
unk            & 0.9        & \colorbox{Yellow}{98.0}      & 1.7         & \colorbox{Yellow}{99.3}        & 1.3   &  \colorbox{Yellow}{99.9}\\   
mask          &   0.8     & \colorbox{Yellow}{86.4}      & 1.5         & \colorbox{Yellow}{94.1}        & 0.5   &  \colorbox{Yellow}{98.2}\\   
att-zero      & 5.1       & \colorbox{Yellow}{78.9}      & 1.9         & \colorbox{Yellow}{58.7}        & 0.6   &  \colorbox{Yellow}{50.8}\\    
\bottomrule
\end{tabular}
\end{center}
\caption{OOD percentage when our counterfactual editor models generate samples, compared to other replacement methods (erase, unk, mask, and att-zero methods). This is the percentage of corrupted examples that are out of the 99th percentile of the NLL of the original sentences in AG-News dataset (lower is better). The scenarios with very high numbers of OODs are highlighted.}
\label{OODs_nnews2}
\end{table*}

\section{} \label{B}
\begin{figure}[t]
\begin{center}

\includegraphics[width=1.02\linewidth]{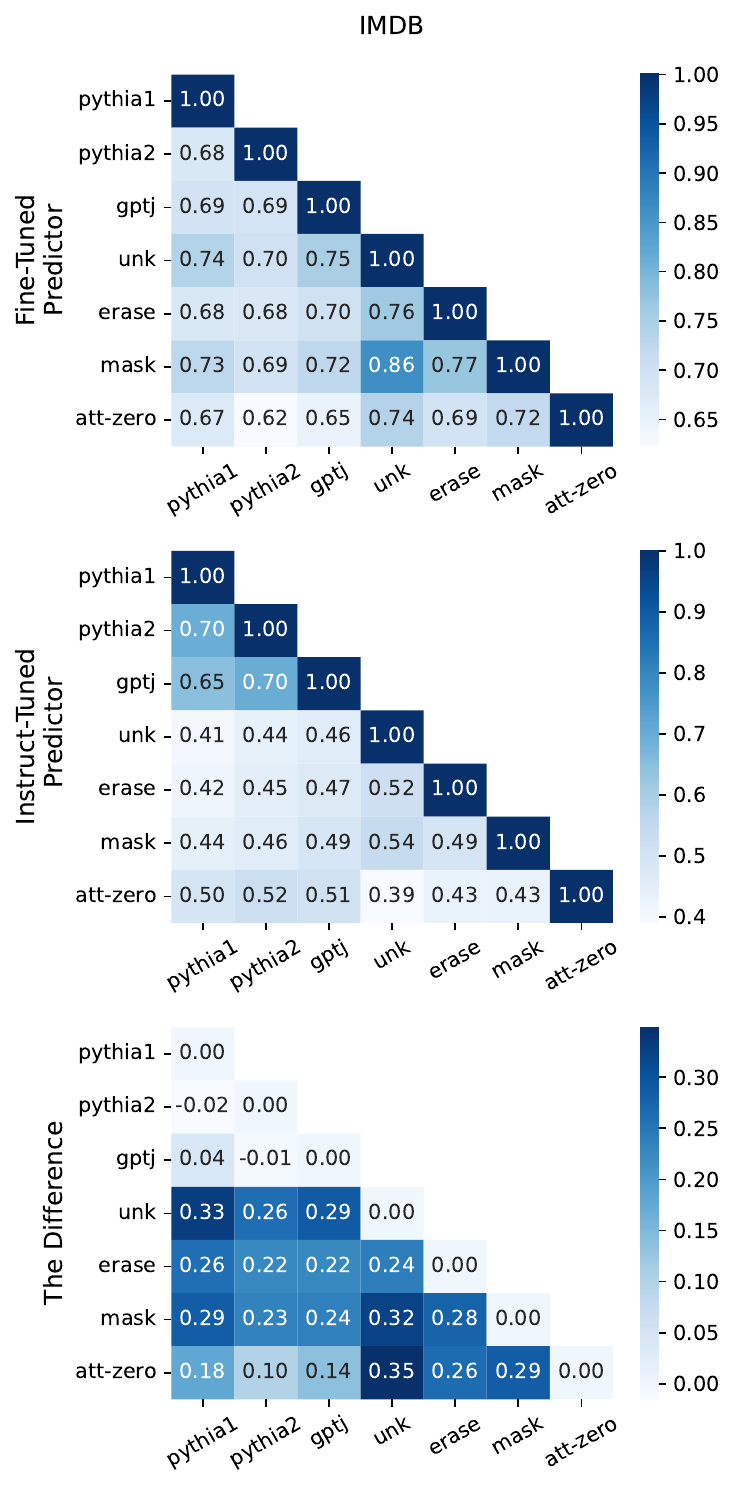}
\end{center}
\caption{The difference}
\label{corr_imdb}
\end{figure}

\begin{figure}[t]
\begin{center}

\includegraphics[width=1.02\linewidth]{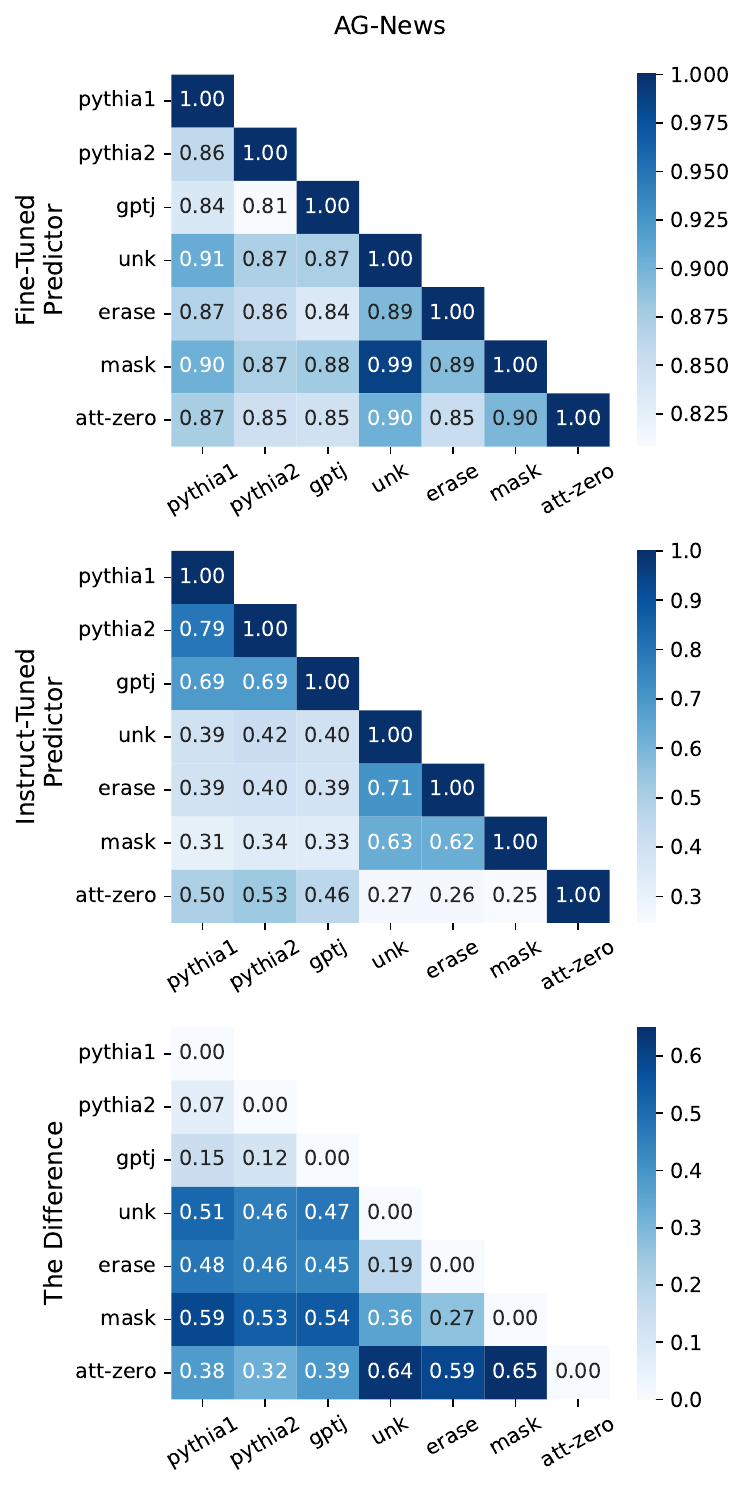}
\end{center}
\caption{The difference}
\label{corr_news}
\end{figure}

Figures \ref{corr_imdb} and \ref{corr_news} show the difference of correlations in IMDB and AG-News datasets respectively.

\section{} \label{C}
Tables \ref{flip-rate-ft} and \ref{flip-rate-it} show flip-rates for fine-tuned and instruct-tuned predictor models respectively.

\begin{table*}[h!]
\begin{center}
\begin{tabular}{m{2cm}||ccc|ccc|ccc}
\toprule
\multirow{2}{0pt}{\bf Attribution method} &\multicolumn{3}{c}{\bf SST-2} &\multicolumn{3}{c}{\bf IMDB} &\multicolumn{3}{c}{\bf AG-News}\\

  &{\bf pythia1} & {\bf pythia2 } &{\bf gptj} &{\bf pythia1} & {\bf pythia2 }&{\bf gptj}&{\bf pythia1} & {\bf pythia2 }&{\bf gptj}\\
\midrule
gradnorm1 & 67.5 & 63.5 &72.5& 78.5 &76.5 &68.8&22.0   &18.5 & 17.0\\
gradnorm2    &66.5      &61.0    & 71.0 &75.5    &80.5 & 67.5 &22.0 &19.5 &17.0\\
gradinp      &32.0      &31.0     &33.0 &54.5    &50.5   & 49.0&21.5   &21.0 &23.0\\
erasure      &56.5      &47.5    & 56.5 &59.0    &61.5 & 55.5& 24.5  &24.0 & 23.0\\
IG           &18.5      &17.5     & 23.5&33.5    &31.5    &41.0 &16.5   &14.5 & 21.0\\
KernelSHAP   &22.0      &17.5    & 22.5 &30.5    &34.0    &31.0&18.0   &13.5 & 18.5\\
Random       &22.5      &18.5     & 23.5&33.5    &38.5    &35.0&18.5   &15.5  &18.5\\
\bottomrule
\end{tabular}
\end{center}
\caption{The mean percentage of success in flipping Gemma-ft's label in 200 examples of evaluation split in SST-2, IMDB, and AG-News datasets (higher is better).}
\label{flip-rate-ft}
\end{table*}
\begin{table*}[h!]
\begin{center}
\begin{tabular}{m{2cm}||ccc|ccc|ccc}
\toprule
\multirow{2}{0pt}{\bf Attribution method} &\multicolumn{3}{c}{\bf SST-2} &\multicolumn{3}{c}{\bf IMDB} &\multicolumn{3}{c}{\bf AG-News}\\

  &{\bf pythia1} & {\bf pythia2 } &{\bf gptj} &{\bf pythia1} & {\bf pythia2 }&{\bf gptj}&{\bf pythia1} & {\bf pythia2 }&{\bf gptj}\\
\midrule
gradnorm1    &34.5   &31.0   & 33.5&37.5    &36.0   & 44.0&12.0       & 10.0 & 18.0\\
gradnorm2    &34.0   &31.5    &33.5&37.0    &35.0   & 43.5&12.0       &10.0 &18.5\\
gradinp      &27.0   &25.0    &28.0&38.5    &36.5   & 62.0&18.5   &18.5 &17.0\\
erasure      &29.5 &26.5 &32.5&29.5    &26.5   & 46.0&17.0       &16.0 &21.5\\
IG           &23.5   &24.5    &23.0&37.0    &38.0   & 61.0&17.5       &18.5  &12.0\\
KernelSHAP   &28.0   &26.0    &21.5&34.0    &36.0   & 59.5 &17.5       &18.0 &19.0\\
Random       &21.0   &23.0    &18.5 &30.0 &30.5 &55.5&20.5   &19.0 &17.0 \\
\bottomrule
\end{tabular}
\end{center}
\caption{The mean percentage of success in flipping Gemma-it's label in 200 examples of evaluation split in SST-2, IMDB, and AG-News datasets (higher is better).}
\label{flip-rate-it}
\end{table*}

\end{document}